\documentclass[10pt, a4paper]{article}

\usepackage{pgfplots}
\usepackage{sansmath} 
\pgfplotsset{compat=1.18}

\usepackage{xcolor}
\usepackage[utf8]{inputenc}
\usepackage{booktabs}
\usepackage{xcolor}
\usepackage{multirow}
\usepackage{colortbl}
\usepackage{rotating}
\usepackage{tabularx}   
\usepackage{adjustbox}  

\usepackage{booktabs}
\usepackage{tabularx}
\usepackage{makecell}

\usepackage[final]{lrec2026} 

\title{ToxSyn-PT: A Synthetic Fine-Grained Dataset of Minority-Targeted Toxic Language in Portuguese}

\name{
\begin{tabular}{c}
Iago A. Brito, Julia S. Dollis, Fernanda B. Färber, Diogo F. C. Silva\\
Arlindo R. Galvão Filho
\end{tabular}
}

\address{Advanced Knowledge Center for Immersive Technologies (AKCIT)\\Federal University of Goiás \\
         \{iagoalves, juliadollis, fernandabufon, diogo\_fernandes\}@discente.ufg.br\\
         arlindogalvao@ufg.br
         }

\abstract{
    The development of robust hate speech detection systems remains limited by the lack of large-scale, fine-grained training data, especially for languages beyond English. Existing corpora typically rely on simplistic toxic and non-toxic labels, and the few that capture hate directed at specific minority groups lack the positive counterexamples required to distinguish genuine hate from mere discussion. In this work, we introduce ToxSyn-PT, the first Portuguese large-scale corpus explicitly designed for multi-label hate speech detection across nine protected minority groups, including the non-toxic counterexamples absent in all other public datasets. Generated via a controllable four-stage pipeline, ToxSyn contains discourse-type annotations to capture rhetorical strategies of toxic/non-toxic language, such as sarcasm, dehumanization, and cultural appreciation. Our experiments reveal a catastrophic, mutual generalization failure compared to existing datasets from social-media domains: models trained on social media struggle to generalize to minority-specific contexts, and vice-versa. This finding indicates they are distinct tasks and exposes summary metrics like Macro F1 can be unreliable indicators of true model behavior, as they completely mask model failure. We publicly release ToxSyn on \href{https://huggingface.co/datasets/AKCIT/ToxSyn-PT}{Hugging Face} to support reproducible research on synthetic data generation and benchmark progress in hate-speech detection for low- and mid-resource languages.
 \\ \newline  }

\begin{document}

\maketitleabstract

{\small \noindent Published in \textit{Proceedings of the Fifteenth Language Resources and Evaluation Conference (LREC 2026)}, pp. 3908--3920. DOI: 10.63317/3ne367tx8hvj.}

\vspace{0.5em}

{\small \noindent\textcolor{red}{\textbf{Warning.} \textit{This paper discusses and contains offensive content.}}}

\section{Introduction}

The task of identifying and mitigating online hate speech is a critical challenge for building safe and inclusive digital spaces, from social media platforms to nascent virtual reality ecosystems \cite{albladi2025hate, weerasinghe2025beyond}. However, progress is limited by severe limitations in available training corpora. Most existing datasets frame toxicity detection as a binary classification problem, lacking the multi-label annotations needed to identify specific protected classes, such as racial, religious, or gender groups \citelanguageresource{vargas-etal-2022-hatebr}. Furthermore, while some English datasets have begun to include both explicit and implicit toxicity, the fundamental aspect of discourse type remains unaddressed in existing corpora, failing to capture crucial contextual features. Consequently, the field is constrained to developing models that may achieve high performance on coarse-grained benchmarks, but lack the necessary robustness and contextual understanding to address the nuanced ways hate speech targets vulnerable communities.

This data-centric challenge is further compounded by a linguistic bias that permeates the field. Despite the global nature of the problem, the vast majority of research and technological progress remains restricted to Anglocentric contexts, relying on models trained with informal English data scraped from a handful of platforms \citelanguageresource{jahan2023systematic}. As a result, these models often fail to generalize across different linguistic and cultural contexts, leading to a significant performance gap, not only restricting the development of universally effective moderation tools but also leaves speakers of most languages, particularly those in low-resource settings, disproportionately vulnerable to online toxicity \cite{kargaran2024mexa}.

These interconnected issues of data granularity and linguistic focus are particularly pronounced in the Portuguese context. Existing datasets suffers from a several limitations, as the few works that attempt to identify hate speech for specific protected groups contains only dozens or hundreds of labeled samples, fail to include benign text about minorities to serve as non-toxic counterexamples, and remains confined in the social media domain, whose linguistic patterns differ markedly from other contexts where toxicity manifests, such as news commentary or transcribed dialogues \citelanguageresource{de2017offensive, leite-etal-2020-toxic, vargas-etal-2022-hatebr}. These omissions and reliance on a single domain fundamentally limit a model's ability to distinguish genuine hate from mere discussion, and prevent the training and evaluation of minority-aware toxicity detection models, highlighting the necessity of better Portuguese classifiers and benchmarks.

In this work, we introduce ToxSyn-PT\footnote{\href{https://huggingface.co/datasets/AKCIT/ToxSyn-PT}{https://huggingface.co/datasets/AKCIT/ToxSyn-PT}}, a large-scale Portuguese dataset addressing minority hate speech detection. Comprising 53,274 LLM-generated samples annotated for toxicity, discourse type, and minority group, ToxSyn covers nine protected communities, including racial, religious, gender, and ability-based categories, all underrepresented in existing resources. Furthermore, the dataset is composed by toxic samples (harmful texts against minority groups), non-toxic samples (positive and neutral sentences referencing minority groups) and neutral samples (texts not referencing any minority groups). To the best of our knowledge, it is the first corpus in Portuguese explicitly designed to support hate-speech detection across multiple protected groups.

We evaluate the effectiveness of ToxSyn by fine-tuning open-source models and testing them on toxic vs. non-toxic classification and minority-targeted toxicity detection. The results show that Portuguese toxicity detection models are strongly domain-dependent, performing well within their training domain but showing substantial degradation in out-of-domain settings. This finding highlights the need for robust, balanced, and minority-aware datasets to achieve reliable generalization across diverse linguistic domains and demographic targets.

Our main contributions are:

\begin{enumerate}
    \item \textbf{Dataset.} We introduce ToxSyn, the first publicly available Portuguese corpus designed to support hate-speech classification across multiple minority targets, comprising over 50K synthetic instances annotated with toxicity, target group, and discourse-type labels.

    \item \textbf{Generation Pipeline.} We present a controllable LLM-based data generation pipeline that enables balancing class distributions, injecting low-frequency expressions, and applying safety constraints.

    \item \textbf{Evaluation.} We demonstrate that toxicity detection in Portuguese is strongly domain-dependent, with model performance degrading sharply when applied to out-of-distribution contexts.
\end{enumerate}

\section{Related Works} \label{sec:related_works}

Synthetic corpora have been explored as a means to augment limited hate speech datasets. ToxiGen \citelanguageresource{hartvigsen-etal-2022-toxigen} combines a classifier-in-the-loop framework with GPT-3 \cite{brown2020language} to adversarially generate over 250,000 examples across 13 demographic targets, yielding measurable improvements in classification performance. ToxiCraft \citelanguageresource{hui2024toxicraft} extends this line of work by using GPT-4 \citep{achiam2023gpt} with structured prompting and self-evaluation to generate text with greater controllability. Despite their technical sophistication, both approaches are constrained to English, which limits their applicability to languages that differ in sociolinguistic norms and expressions of toxicity.

In the context of Portuguese, HateBR \citelanguageresource{vargas-etal-2022-hatebr} provides a multilabel dataset of 7,000 Instagram comments, annotated for binary offensiveness, severity (in three levels), and nine hate speech categories. Although the annotation schema is comprehensive, the category distribution is highly uneven: only 727 category labels are assigned in total, with 496 corresponding to partyism and fewer than 100 for any other class (e.g., just two antisemitic samples and a single example of xenophobic text). This imbalance limits the dataset's utility for training models with generalization capacity.

Several additional Portuguese-language corpora have been proposed, but often suffer from limited scale and class coverage. OFFCOMBR-3 \citelanguageresource{de2017offensive} includes 1,250 comments annotated across six categories by three annotators, but only 19.5\% of the samples are toxic and most categories have very few labeled instances marking harmful content against a minority group. ToLD-BR \citelanguageresource{leite-etal-2020-toxic} provides annotations for four minority classes, although it includes less than 30 examples per class. OLID-BR \citelanguageresource{trajano2024olid} improves on label balance by annotating between 92 and 461 samples per category across five target classes. Therefore, the dataset size remains limited for training deep learning models without additional data augmentation.

\begin{figure*}[t]
    \centering
    \includegraphics[width=\textwidth]{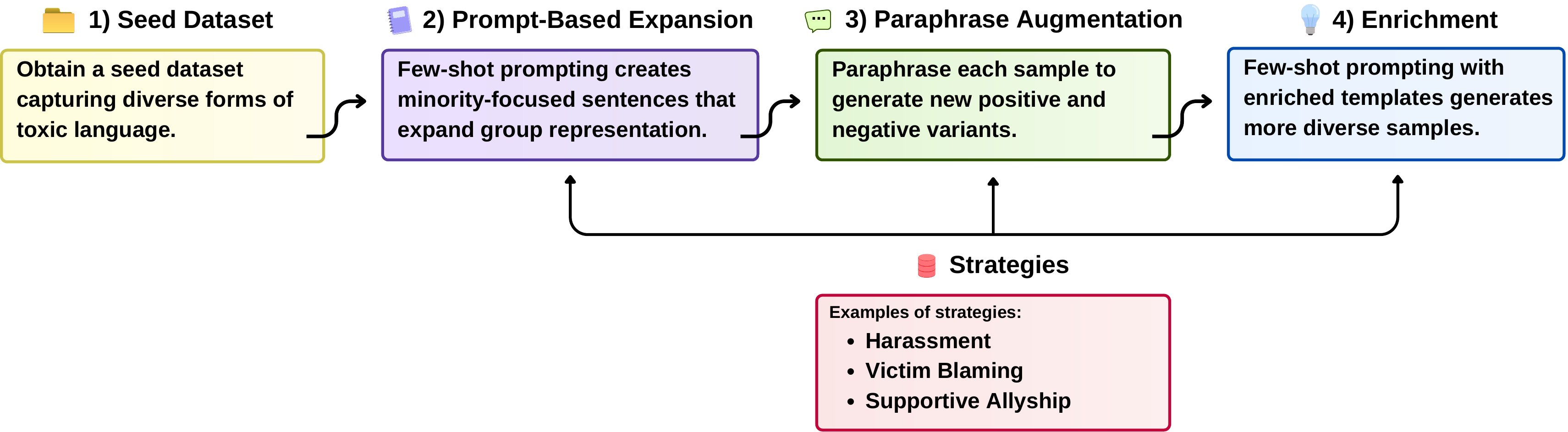}
    \caption{Overview of ToxSyn generation pipeline.}
    \label{fig:proposed_pipeline}
\end{figure*}

TuPy-E \citelanguageresource{oliveira2023tupy} merges three annotated Portuguese corpora \citelanguageresource{fortuna-etal-2019-hierarchically, leite-etal-2020-toxic, vargas-etal-2022-hatebr} with the TuPy dataset into a unified collection of 43,668 social-media comments. Among them, 11,547 are labeled as toxic, and only 3,327 carry annotations indicating offense toward a protected minority\footnote{Although TuPy-E contains \textit{Political} hate category, we disconsider it since this class do not offend any minority group.}. Although \mbox{TuPy-E's} multilabel schema spans nine categories (e.g., ageism, aporophobia, capacitism, LGBT-phobia), its utility is undermined by severe class imbalance: some categories are represented by fewer than one hundred instances (e.g. ageism and aporophobia has 57 and 66 examples, respectively), whereas there is categories such as misogyny  that comprises 1,675 samples. This extreme sparsity degrades the performance of encoder-based multilabel classifiers in the original study, highlighting the urgent need for larger, more evenly distributed resources to train and evaluate robust hate-speech detection models in Portuguese.

Additionally, existing Portuguese corpora exhibit an additional blind spot: they annotate only hostile mentions of protected groups and omit neutral or supportive references. Consequently, a sentence that praises Black resilience or affirms LGBTQIA+ rights is either absent or, at best, indistinguishable from non-minority content. This omission prevents classifiers from learning the difference between the absence of hate and the presence of endorsement, limiting applications such as allyship detection and fine-grained content moderation in both formal and informal domains. Filling this gap is essential for building context-aware systems that recognise not just toxicity, but also positive engagement with minority communities.

\section{ToxSyn} \label{sec:method}

Our approach to construct ToxSyn emphasizes controlled representation of protected groups and linguistic diversity across both toxic and non-toxic expressions. As illustrated in Figure~\ref{fig:proposed_pipeline}, the dataset is produced through a structured four-stage generation pipeline, designed to ensure broad coverage of different target communities while capturing a range of implicit and explicit toxic language. To further enhance generalization and reduce potential overfitting to specific groups, we also incorporate neutral content unrelated to minority identity, increasing topical variety across the dataset. Detailed descriptions of each stage of the pipeline are provided in the following subsections.

\subsection{Seed Dataset}

We hypothesize that exposing a language model to a rich spectrum of perspectives and expressions of toxicity for a compact set of minority targets enables better generalization to additional, unseen communities. Guided by this principle, we constructed a focused seed dataset centered on two targets, Black people and women. Selecting a small set of groups allowed us to pursue depth over breadth, systematically exploring intra-group variation in how hostility and positivity are expressed while keeping the seed compact and semantically coherent for use as in-context demonstrations during later prompt-based expansion.

Concretely, the seed comprises 40 toxic and 40 non-toxic human written examples for each target, finalizing with 160 sentences in total. The samples were crafted to vary across multiple axes (i.e. tone, topical content, and syntactic structure), maximizing intra-group diversity and providing clear positive and counterfactual instances that help the model distinguish targeted hostility from benign references. These high-quality seed examples form the foundation for our subsequent controlled LLM generation and augmentation stages. Examples of samples present in the seed dataset are available in Appendix \ref{app:ap_seed_samples}.

\subsection{Prompt-Based Expansion} \label{sec:expansion}

In this stage, we employ GPT-4o Mini \citep{achiam2023gpt} in a few-shot prompting configuration to generate new samples conditioned on minority identity. The model is prompted with representative seed examples and instructed to produce linguistically diverse toxic and non-toxic statements for one of nine target communities: Black people, Jews, Muslims, Indigenous Brazilians, women, LGBTQIA$+$ individuals, elderly people, and people with disabilities. These categories were selected to achieve both broad demographic coverage and equitable representation of groups that are typically under-sampled or totally absent from existing Portuguese hate-speech datasets.

To ensure wide topical and stylistic coverage across all minorities, we developed 26 distinct discourse type templates (14 non-toxic, 12 toxic) that systematically vary the ideological frame and tone of the outputs. For example, the \emph{Real Problems Minimization} discourse guides the model to downplay the hardships of marginalized communities, whereas the non-toxic template \emph{Problem Acknowledgment} promotes constructive framing of societal challenges. The full list of discourses are shown in Appendix~\ref{app:ap_generation_strategies}. To further enhance textual diversity, each generation prompt also includes a randomized length constraint (short, medium, or long).

We guarantee both representational balance and conceptual robustness by generating an evenly distributed dataset across both demographic groups and toxicity labels. This deliberate sampling design not only prevents asymmetric group representation but also supports fairer evaluation of language models across demographic dimensions. The resulting corpus comprises 10,790 unique sentences that combine linguistic richness with demographic diversity, forming a solid foundation for subsequent stages of large-scale synthetic expansion.

\subsection{Paraphrase Augmentation}

Each sentence produced in the expansion stage was paraphrased twice, once through a toxic and once through a non-toxic transformation, enriching lexical and pragmatic variation. These transformations were randomly sampled from a new predefined discourse strategy pool, introducing diverse rhetorical and stylistic shifts that reflect how toxicity and neutrality manifest in natural discourse. Toxic transformations simulate linguistic mechanisms such as victim-blaming, ambiguous framing, appeal to authority, and hyperbolic exaggeration, while non-toxic transformations leverage strategies like questioning, contrastive emphasis, nuanced ambiguity, and positive negation (see Appendix~\ref{app:ap_rewriting} for more detailed descriptions). This procedure enhances linguistic heterogeneity and pragmatic realism, providing the model with broader exposure to the subtle ways toxicity and neutrality are expressed.

Because our transformations are applied symmetrically (i.e., non-toxic rewriting patterns are applied to originally toxic inputs and toxic patterns to originally non-toxic inputs), we frequently obtain instances that are effectively detoxified but retain residual pragmatic or referential ambiguity regarding intent and target, as well as instances where originally non-toxic inputs become ambiguous toxic samples. These ambiguous examples increase the corpus realism by reflecting positive text but with terms usually associated with natural hostility discourse, serving as valuable edge cases as they reduce reliance on lexical heuristics by encouraging models to exploit contextual and pragmatic cues, providing hard examples useful for both training and evaluation phases.

Furthermore, our paraphrasing process enabled systematic swapping of the original minority target across the nine predefined groups. However, LLMs outputs occasionally hallucinated unsupported targets or produced non-standard surface forms. To enforce consistency, we applied a regular-expression based normalization that maps variant forms to our nine canonical labels (e.g., all ``immigrant from [...]'' variants were mapped to ``immigrant''). This normalization resulted in the discarding of approximately 8\% of generated sentences that lacked a valid mapping.

After normalization and filtering, this stage produced 20,799 additional paraphrases. When combined with the previous samples, the corpus achieves 14,521 toxic and 17,068 non-toxic instances, a total of 31,589 exemplars. These procedures substantially broaden ToxSyn's stylistic and rhetorical diversity while preserving precise minority-target annotations.

\subsection{Enrichment}

In the final generation round we repeat the few-shot procedure from Section~\ref{sec:expansion}, but draw our in-context examples from the 31,589 sentences produced rather than the 160 examples from the original seed dataset. By exposing GPT-4o Mini \citep{achiam2023gpt} to this richer demonstration pool, already populated with a spectrum of explicit and implicit rhetoric, we encourage the model to imitate a broader range of discourse styles while preserving the label fidelity established earlier.

To further broaden the corpus's coverage of implicit hate speech, we extend the template catalogue from 26 to 28 by adding two new toxic discourses: \emph{Ambiguous Prejudice} (double-meaning language to convey prejudice in an implicit but perceptible way) and \emph{Justification Prejudice} (frames discriminatory attitudes as ostensibly reasonable, presenting bias as a matter of common sense, economic necessity, or cultural tradition). Although earlier stages included implicit hate, these two templates diversify the rhetorical tactics used to conceal prejudice, producing harder-to-classify examples that increase the dataset's linguistic complexity.

The generation process produced a minority-targeted corpus of 50,074 sentences, including 24,707 toxic and 25,367 non-toxic examples with automatically generated labels. A portion of this dataset was subsequently reviewed by human annotators to validate label quality. The resulting corpus spans a range of hateful expressions, from explicit slurs to more subtle forms of prejudice, alongside neutral or benign statements that provide contrast for hate speech detection research.

\begin{table}[h]
    \centering
    \small
    \adjustbox{width=0.48\textwidth}{
    
    \begin{tabular}{lcccc}
        \toprule
        \textbf{Group} & \textbf{Train Count} & \textbf{Test Count} & \textbf{Total} \\
        \midrule
        \textbf{Black} &  &  &  \\
        \hspace{1em}Toxic & 2,688 & 308  & 2,996 \\
        \hspace{1em}Non-toxic & 2,714 &  299  & 3,013 \\
        \hspace{1em}Count & 5,402 & 607 & 6,009 \\
        
        \textbf{Women} &  &  &  \\
        \hspace{1em}Toxic & 2,703 & 320 & 3,023 \\
        \hspace{1em}Non-toxic & 2,307 & 248  & 2,555 \\
        \hspace{1em}Count & 5,010 & 568 & 5,578\\
        
        \textbf{LGBTQIA+} &  &  &  \\
        \hspace{1em}Toxic & 2,653 & 303 & 2,956 \\
        \hspace{1em}Non-toxic & 2,668 & 285 & 2,953 \\
        \hspace{1em}Count & 5,321 & 588 & 5,909 \\
        
        \textbf{Native Brazilian} &  &  &  \\
        \hspace{1em}Toxic & 2,615 & 293 & 2,908 \\
        \hspace{1em}Non-toxic & 2,659 & 292 & 2,951 \\
        \hspace{1em}Count & 5,274 & 585 & 5,859\\
        
        \textbf{Muslim} &  &  &  \\
        \hspace{1em}Toxic & 2,601 & 287 & 2,888 \\
        \hspace{1em}Non-toxic & 2,592 & 285 & 2,877 \\
        \hspace{1em}Count & 5,193 & 572 & 5,765\\
        
        \textbf{Jewish} &  &  &  \\
        \hspace{1em}Toxic & 2,514 &  289 & 2,803 \\
        \hspace{1em}Non-toxic & 2,569 & 269 & 2,838 \\
        \hspace{1em}Count & 5,083 & 558 & 5,641\\
        
        \textbf{Elderly} &  &  &  \\
        \hspace{1em}Toxic & 2,039 & 237  & 2,276 \\
        \hspace{1em}Non-toxic & 2,433 & 257  & 2,690 \\
        \hspace{1em}Count & 4,472 & 494  & 4,966\\
        
        \textbf{Disabled People}&  &  &  \\
        \hspace{1em}Toxic & 2,400 & 270  & 2,670 \\
        \hspace{1em}Non-toxic & 2,489 & 269  & 2,758 \\
        \hspace{1em}Count & 4,889 & 539  & 5,428\\
        
        \textbf{Immigrants} &  &  &  \\
        \hspace{1em}Toxic & 2,022 & 236  & 2,258 \\
        \hspace{1em}Non-toxic & 2,400 & 249  & 2,649 \\
        \hspace{1em}Count & 4,422 & 485  & 4,907\\
        
        \textbf{Neutral} &  &  &  \\
        \hspace{1em}Toxic & 0 & 0 & 0 \\
        \hspace{1em}Non-toxic & 3,000 & 212  & 3,212 \\
        \midrule
        \textbf{Total} & \textbf{48,066} & \textbf{5,208} & \textbf{53.274} \\
        \bottomrule
    \end{tabular}
    }
    \caption{Final distribution of ToxSyn-PT across groups. Labels in the test set were validated by human annotators, whereas train set labels were automatically generated during the dataset generation process.}
    \label{tab:group_analysis}
\end{table}

\subsection{Neutral Samples}

To improve generalization to content where demographic cues are absent, we design a two-phase neutral-text augmentation module. In Phase 1, we curated 60 fully neutral handwritten sentences and expand each via a five-shot prompting routine using four randomly chosen domain templates (conversational, news, policy, and academic). Each five-shot input return five new samples as output, resulting in 300 domain-varied variants.

In Phase 2, we further diversify those 300 sentences through 20 distinct style-transformation prompts, ranging from social-media vernacular to formal editorials, resulting on 6,000 neutral candidates. To maintain emphasis on minority-targeted hate speech, we stratify this pool by seed, domain, and style, then uniformly sample 3,200 sentences for integration. This procedure injects rich topical and stylistic diversity while preserving the dataset's core focus on protected-group samples.

\subsection{Human Annotation}

To create a high-quality, human-verified benchmark, we constructed a test set of 5,208 examples. This set was originally composed of 200 neutral samples and a 10\% sample of the generated minority-targeted data (5,008 examples). To ensure comprehensive representation of our fine-grained categories, this sample was stratified across the \textit{Toxic Label}, \textit{Minority}, and \textit{Discourse Type} features.

A red team of three native Portuguese-speaking annotators labeled each sample as toxic or non-toxic and identified the targeted minority group. To ensure an unbiased gold standard and prevent the cognitive bias introduced by pre-annotation \cite{beck2025bias}, annotators were kept blind to the machine labels, annotating from scratch all 5,208 samples from the raw text. This approach served a dual purpose: 1) to efficiently produce a gold-standard test set and 2) to directly quantify the quality of our synthetic generation pipeline.

Annotators were instructed to label a sentence as toxic if it contained toxic content without explicit criticism or rejection of the statement. For instance, "my brother thinks black people are inferior" should be labeled toxic, whereas "my brother thinks black people are inferior, but this view is wrong because all people are equal" should be labeled non-toxic. Prior to annotation, participants received detailed guidelines and content warnings about the offensive nature of the data, along with the option to skip items or withdraw at any time without penalty.

The results of this validation strongly affirm the high fidelity of our synthetic data. The human annotators modified around 6\% of the primary toxicity labels and fewer than 1\% of the specific minority target labels. This high level of agreement between the generated labels and the human judgment demonstrates the robustness and accuracy of our controllable pipeline.

\begin{table}[t]
\centering
\adjustbox{width=0.48\textwidth}{
\begin{tabular}{llcc}
    \toprule
    \textbf{Datasets} & \textbf{Source} & \textbf{Groups} & \textbf{Target}\\
    \midrule
    OffcomBR-3    & News comments   &   6   &  19    \\
    ToLD‑BR       & Twitter         &   4   &  124    \\
    HateBR        & Instagram       &   9   &  727    \\
    OlidBR        & Social media    &   5   &  1,390  \\
    Tupy-E        & Social media    &   9   &  3,742  \\
    \midrule
    \textbf{ToxSyn (ours)} & \textbf{Controlled generation} & \textbf{9} & \textbf{50,062} \\
    \bottomrule
\end{tabular}}
\caption{Comparison of hate‑speech datasets in Portuguese by source, number of target groups, and number of samples targeting a minority group.}
\label{tab:comparison}
\end{table}


\subsection{Final Dataset}

The final ToxSyn corpus distribution is shown in Table \ref{tab:group_analysis}. It comprises 53,274 synthetic sentences, including 24,778 toxic examples and 25,284 non-toxic sentences referencing minority groups, and 3,212 non-toxic neutral samples, resulting in a near-balanced split (47\% toxic vs. 53\% non-toxic), with each of the nine protected groups represented by at least 4,907 instances and at most 6,009 samples. Every entry contains annotations for toxic label, minority group, and discourse type, supporting both supervised model training and corpus-level analyses. In Table \ref{tab:comparison}, we contrast ToxSyn's source, group coverage, and sample counts with existing Portuguese datasets. Representative examples illustrating diversity in target, toxicity, and discourse style are provided in Table \ref{tab:pt_ToxSyn_examples}.


\begin{table*}[t]
  \centering
  \adjustbox{width=1\textwidth}{
  \begin{tabular}{@{}p{0.85\textwidth}ccc@{}}
    \toprule
    \textbf{Comment} & \textbf{Offensive} & \textbf{Minority} & \textbf{Discourse Type} \\
    \midrule
    Os muçulmanos frequentemente clamam por respeito, mas será que sua história de conflitos não complica essa busca por aceitação?\\[-2pt]
    {\color{blue}\textbf{Translation:}} Muslims often call for respect, but doesn't their history of conflicts complicate this quest for acceptance? 
      & Yes  & Muslims & Ambiguous Prejudice     \\
    \addlinespace
    \midrule
    Se as mulheres realmente quisessem ajudar, ao invés de criarem mais divisão, poderiam ter contribuído positivamente para os movimentos sociais.\\[-2pt]
    {\color{blue}\textbf{Translation:}} If women truly wanted to help, instead of creating more division, they could have contributed positively to social movements.
      & Yes & Women & Victim Blaming   \\
    \addlinespace
    \midrule
    Não se pode ignorar que a comunidade judaica é frequentemente reconhecida por sua resiliência, mesmo em situações desafiadoras.\\[-2pt]
    {\color{blue}\textbf{Translation:}} It cannot be ignored that the Jewish community is often recognized for its resilience, even in challenging situations. 
      & No & Jewish & Positive Negation \\
    \addlinespace
    \midrule
    Às vezes, dá até um nó na cabeça ver a rapidez que as fofocas rolam nas redes sociais, mas é massa perceber como isso mexe com a cabeça da galera.\\[-2pt]
    {\color{blue}\textbf{Translation:}} Sometimes it's mind-boggling how fast gossip spreads on social media, but it's cool to see how it really messes with people's minds.
      & No & Neutral & -   \\
    \bottomrule
  \end{tabular}
  }
  \caption{Samples from ToxSyn labeled with Hate-Speech, Minority and Discourse Type. In regionalism discourse, it is possible to see some Brazilian Portuguese slugs.}
  \label{tab:pt_ToxSyn_examples}
\end{table*}


\section{Experiments} \label{sec:experiments}

We evaluate the effectiveness of the ToxSyn dataset under two classification settings: (1) multi-domain classification, which determines whether a given text is toxic or non-toxic in both general and minority targeted benchmarks, and (2) group-specific classification, where a separate model is trained for each protected group to determine whether a text is toxic toward that group.

Given the lack of Portuguese benchmarks with detailed multi-label annotations, we translated the human-annotated portion of ToxiGen \citep{hartvigsen-etal-2022-toxigen} into Portuguese using GPT-4 \citep{achiam2023gpt}. These experiments aim to assess the impact of synthetic data generated by our pipeline across multiple evaluation settings and model capacities.

\subsection{Experimental Setup}

\paragraph{Implementation.}
All experiments were conducted on a single NVIDIA RTX 4090 GPU using the Hugging Face Transformers library\footnote{https://huggingface.co/}. We used a batch size of 32, a learning rate of 5e-5 with linear scheduling, and no weight decay. Models were trained for 3 epochs.

\paragraph{Model.}
For both multi-domain and group-specifi tasks, we fine-tuned the BERTimbau base model \citep{souza2020bertimbau}, which has demonstrated strong performance in prior Portuguese NLP benchmarks \citep{da2024toxic}.

\paragraph{Classification.}
In the multi-domain classification setup, the model was trained on all training examples and evaluated on both social media and synthetic data domains. For the group-specific classification setting, we constructed a separate training set for each protected group. All toxic samples targeting the group were retained as positive instances, while 3,000 counterexamples were randomly sampled as negatives: 1,000 non-toxic samples from the same group, 1,000 toxic samples targeting other groups, and 1,000 neutral samples.

\subsection{Evaluation Benchmarks}

Existing Portuguese hate-speech corpora are insufficient for robust target-group evaluation because they contain only tens to hundreds of examples per protected class, which is far too few for reliable performance assessment. To mitigate this, we concatenated public datasets that contain group annotations (HateBR \citelanguageresource{vargas-etal-2022-hatebr}, OFFCOMBR-3 \citelanguageresource{de2017offensive}, OLID-BR \citelanguageresource{trajano2024olid}, and TuPy-E \citelanguageresource{oliveira2023tupy}), merging related groups to increase the number of test samples. For example, all nationality-based groups were grouped under \textit{immigrants}, while Antisemitic and Islamophobic references were subsumed into \textit{religious intolerance} given the limited number of samples available for each category. The merged dataset contains 6,002 samples, of which 2,318 share at least one protected-group label with ToxSyn. For simplicity, we mention this set as Portuguese Merged in the rest of the paper.

To improve minority coverage for evaluation, we utilize GPT-4o Mini \citep{achiam2023gpt} to translate the human-annotated portion of ToxiGen \citep{hartvigsen-etal-2022-toxigen}. This set contains multi-label examples for six protected groups that are also present in ToxSyn. The resulting dataset contains 8,960 samples of both toxic and non-toxic text mentioning specific groups, of which 6,166 samples contains minority groups label that intersects with ToxSyn.

Although Portuguese Merged set constitutes the largest multi-label test set for hate speech in Portuguese, it is constrained by its informal social-media register and by only marking protected groups when content is toxic; neutral or supportive references remain unlabeled, and no existing corpus covers enough minority-targeted examples for training. The translated ToxiGen dataset mitigates some of these gaps by supplying formal, multi-label examples, but it introduces unnaturalized phrasing and culturally mismatched slurs (for instance, the English "watermelon" stereotype against Black people does not translate meaningfully in Portuguese).

\subsection{Multi-Domain Classification}

We assess cross-domain generalization by fine-tuning BERTimbau on the full training sets of ToLD-BR, HateBR, OLID-BR, and our ToxSyn dataset, subsequently evaluating each model against the test sets of all other benchmarks. To effectively capture performance across these diverse distributions, we employ the Macro F1-score to account for class imbalance, alongside Toxic-Class Recall to specifically prioritize the model's sensitivity to harmful instances. As detailed in Table~\ref{tab:binary-results-updated}, our evaluation reveals distinct performance patterns where models excel within their specific domains but struggle to generalize across the boundary between general social media and minority-focused content.

Initial results demonstrate that models perform robustly when tested on data that matches their training domain or rhetorical style. Models trained on social media (ToLD-BR, HateBR, OLID-BR) achieve strong in-domain recall (ranging from 0.60 to 0.95), indicating they effectively capture the explicit nature of online harassment. Similarly, the model trained on ToxSyn corpus performs well on its own test while effectively generalizes to the translated ToxiGen dataset, achieving a strong 0.69 F1-score and 0.77 Toxic-Class Recall. This transferability suggests that while ToxSyn is synthetic, it successfully encodes the underlying, structural patterns of group-targeted hate, allowing it to identify toxicity in other minority-focused contexts that share a similar objective.

However, moving outside these specific domains reveals a catastrophic, mutual generalization failure. When models trained on general social media are tested on minority-focused data, they appear nearly blind to the toxicity. For instance, the ToLD-BR model identifies only 10 of the 2,472 toxic samples in the ToxSyn test set, and even the strongest social media model (trained on OLID-BR dataset) captures only 20\% of toxic samples in ToxSyn and 19\% in ToxiGen. This failure is reciprocal: the ToxSyn model also struggles with out-of-domain corpora, achieving a toxic-class recall of only 0.17 on ToLD-BR.

These findings indicate that social media toxicity, which is often explicit and impulsive, contrasts with minority-targeted hate, which tends to be more structural or implicit, resulting in fundamentally different linguistic signatures. Consequently, models overfit to their specific rhetorical contexts, highlighting a massive reliability gap and the critical need for diverse, minority-aware datasets to build truly robust classifiers.

\begin{table}[h!]
\centering
\adjustbox{width=\columnwidth}{
    \begin{tabular}{llcccc} 
    \toprule
    & \multirow{2}{*}{\textbf{Test Data}} & \multicolumn{4}{c}{\textbf{Finetune Data}} \\
    \cmidrule(lr){3-6}
    & & \textbf{ToLD-BR} & \textbf{HateBR} & \textbf{OLID-BR} & \textbf{ToxSyn (Ours)} \\
    \midrule
    \multirow{5}{*}{\parbox[c][2.1cm][c]{.5cm}{\centering\rotatebox{90}{\textbf{F1-Score}}}}
    & ToLD‑BR         & 0.79 & 0.69 & \textbf{0.80} & 0.46 \\
    & HateBR          & 0.77 & \textbf{0.91} & 0.63 & 0.46 \\
    & OLID‑BR         & 0.60 & 0.62 & \textbf{0.69} & 0.42 \\
    & ToxiGen*        & 0.41 & 0.46 & 0.48 & \textbf{0.69} \\
    & ToxSyn (Ours)*  & 0.35 & 0.36 & 0.37 & \textbf{0.94} \\
    
    \midrule
    \multirow{5}{*}{\parbox[c][2.1cm][c]{.5cm}{\centering\rotatebox{90}{\textbf{Recall}}}}
    & ToLD‑BR & 0.78 & 0.78 & \textbf{0.95} & 0.17 \\
    & HateBR  & 0.60 & 0.89 & \textbf{0.91} & 0.51 \\
    & OLID‑BR & 0.70 & 0.88 & \textbf{0.94} & 0.47 \\
    & ToxiGen*  & 0.09 & 0.17 & 0.19 & \textbf{0.77} \\
    & ToxSyn (Ours)*  & 0.00 & 0.18 & 0.20 & \textbf{0.94} \\

    \bottomrule
    \end{tabular}
}

\caption{Cross-domain generalization results. We report Macro F1-score and Recall for the Toxic class. Datasets marked with * belong to the domain of group-targeted data.}
\label{tab:binary-results-updated}
\end{table}

\subsection{Group-Specific Classification}

We next evaluated our ToxSyn-trained model's ability to perform fine-grained, multi-label classification of which protected group is being targeted, using the Portuguese Merged and the translated subset of ToxiGen as evaluation sets. A critical challenge in this evaluation is that existing Portuguese datasets contain only toxic instances for each minority group, lacking any neutral or benign counterexamples. This data structure makes precision-based metrics like F1-score mathematically unreliable, as there are no true negatives for the group-specific task. We therefore report Macro Recall, which properly measures the model's sensitivity to detecting targeted hate. In addition, the lack of suitable public Portuguese resources make it impossible to train comparable minority-aware baselines, highlighting the novelty of our work.

As shown in Table~\ref{tab:multilabel-results}, the resulting model achieved a strong average macro recall of 0.70 on the translated ToxiGen set, reflecting the close conceptual alignment between the two minority-focused resources. The model achieves a moderate 0.64 macro recall on Portuguese Merged, demonstrating a solid capability to transfer its knowledge to the noisier, out-of-domain social media context. However, the performance on this set varies significantly across groups, often correlating with data scarcity (e.g., 0.80 recall for "Religious" with 68 samples vs. 0.54 for "Elderly" with 44 samples). These results confirm that ToxSyn can serve as a foundational resource for training robust, minority-aware models in Portuguese, while also highlighting the need for more balanced, multi-group data collection.

\begin{table}[t]
    \tiny
    \centering
    \adjustbox{width=0.48\textwidth}{
    
    \begin{tabular}{llcc}
    \toprule
      & \textbf{Group} & \textbf{Macro-Recall} & \textbf{Support}\\
    \midrule
    \multirow{7}{*}{{\centering\rotatebox{90}{\textbf{PT Merged}}}}
    
      & Black  & 0.63 & 230\\
      & Women  & 0.63 & 961\\
      & LGBTQIA+  &  0.64 & 734\\
      & Religious &  0.80& 68\\
      & Elderly  &  0.54 & 44\\
      & Immigrants  & 0.59 & 281\\
      & \textbf{Average} & \textbf{0.64}  & \textbf{-}\\
      \midrule
    \multirow{7}{*}{{\centering\rotatebox{90}{\textbf{ToxiGen}}}}
      & Black & 0.72 & 713\\
      & Women &  0.74 & 717\\
      & LGBTQIA+ & 0.67 & 714\\
      & Muslim & 0.69 & 688\\
      & Jewish & 0.76 & 688\\
      & Immigrants & 0.60 & 2,646\\
      & \textbf{Average} & \textbf{0.70}  & \textbf{-}\\
    \bottomrule
    \end{tabular}
    }
    \caption{Group‐specific classification performance per target group on ToxiGen Translated and Portuguese Merged.}
    \label{tab:multilabel-results}
\end{table}

\section{Discussion} \label{sec:discussion}

Our experiments reveal a profound domain dependency in Portuguese toxicity detection. The cross-domain evaluation demonstrates a mutual and catastrophic generalization failure, in which models trained on general social media corpora are incapable of detecting minority-targeted hate and, conversely, our ToxSyn-trained model, while highly effective in its own domain, also fails to generalize to general-domain datasets. This mutual failure strongly suggests that toxicity is not a monolithic concept. The rhetorical and contextual patterns of minority-targeted hate are fundamentally different from the lexical cues of social-media insults. This finding exposes a critical methodological risk in toxicity evaluation, revealing that Macro F1 scores can be dangerously deceptive by masking a model's complete failure to perform its primary task.

Given this domain-specificity, the primary contribution of ToxSyn is not as a general-purpose solver, but as the first resource to enable a previously impossible task in Portuguese: building and evaluating fine-grained, minority-aware models. Our group-specific experiments underscored this, as the lack of non-toxic counterexamples in all other public corpora made it impossible to even train comparable baselines.

ToxSyn's success in this task, and its ability to generalize to other minority-focused dataset, is a direct result of our controllable generation pipeline. By systematically creating fine-grained annotations (discourse types, target groups) and, crucially, their corresponding detoxified counterexamples, we compel the model to learn the deeper contextual features that distinguish genuine harm from benign discussion of identity, rather than brittle, surface-level heuristics.

\section{Conclusion} \label{sec:conclusion}

In this paper, we introduced ToxSyn, a large-scale, fine-grained synthetic dataset that fundamentally addresses a critical resource gap in Portuguese. It is the first Portuguese corpus to enable the fine-grained classification of hate against specific minority groups, a task previously impossible due to the critical absence of non-toxic counterexamples in all other public data. Our controllable four-stage pipeline was designed to systematically generate these balanced and nuanced instances, providing a unique resource for the community.

The empirical investigation using ToxSyn revealed that toxicity detection is not a monolithic problem. We demonstrated a catastrophic, mutual generalization failure between general-domain and minority-targeted hate speech, proving that even within the general toxicity domain, there are multiple manifests of harmful content and, therefore, multiple sub-domains. This finding also serves as a crucial methodological warning: our results show that summary metrics like Macro F1 are dangerously deceptive, as they can completely mask a model's total failure to detect the target toxic class. We thus argue that future benchmarks must incorporate more granular metrics. Our generation pipeline provides a generalizable protocol for creating such resources, and we release ToxSyn as a new benchmark to stimulate more rigorous, context-aware evaluation in the pursuit of genuinely equitable online safety.

\section{Future Work} \label{sec:future_works}

Our work on ToxSyn opens several promising avenues for future research. First, our four-stage generation pipeline serves as a generalizable framework. A clear next step is to adapt this methodology to other low-resource languages that face similar data scarcity, enabling the creation of fine-grained, balanced hate speech corpora globally. Second, our experiments confirmed a severe, mutual generalization failure between social-media and minority-focused domains. Future work should focus on bridging this domain gap. This could involve domain creating hybrid datasets that mix different domains to train a single, robust classifier that operates effectively across diverse linguistic contexts. Third, our dataset contains rich discourse-type labels (e.g., Ambiguous Prejudice, Justification Prejudice). While we used these for generation and stratification, they remain an untapped resource. Training models to explicitly classify these rhetorical strategies, moving beyond if a text is toxic to how and why it is toxic, would enrich the research in the field. Finally, while ToxSyn is validated against human annotation, it remains a synthetic resource. A crucial next step is to address this gap by collecting in-the-wild, human-authored data and annotating it using ToxSyn's detailed, multi-label schema.

\section{Limitations} \label{sec:limitations}

While ToxSyn substantially advances Portuguese hate-speech resources, it carries several important caveats. First, each sentence is annotated with a single target group, precluding the analysis of intersectional or multi-target cases; overlapping abuses (e.g., simultaneously sexist and racist language) thus remain unmodeled. Second, the corpus lacks severity tiers (e.g., mild versus severe toxicity), which limits studies requiring fine-grained harm assessment. Third, our four-stage generation pipeline relies on a fixed set of prompts and normalization rules. Although this ensures control and class balance, it may constrain the emergence of novel or context-specific hate-speech patterns not captured in our templates.

Moreover, as a fully synthetic dataset, ToxSyn inherits potential biases from the underlying LLM, including repetitive phrasing and distributional artifacts that may diverge from authentic human language. Finally, while our human-in-the-loop validation of the 5,208 sample test set allowed us to quantify the pipeline's high fidelity, the remaining training samples did not undergo this same rigorous human review and may contain a similar level of labeling noise.

\section{Ethics Statement}
The use of generative models to create synthetic toxic data must be approached with great caution to prevent harmful applications, such as deliberately offending minority groups or training language models to produce hate speech. Nevertheless, responsible use of this methodology can significantly enhance online communication by improving the identification of toxic content and specifying the particular minority groups being targeted. To ensure ethical deployment, applications that intend to use ToxSyn should be validated by a under multidisciplinary team, providing critical guidance on responsible usage and preventive measures for potential misuse.

\section{Acknowledgments}
This work has been fully/partially funded by the project Research and Development of Algorithms for Construction of Digital Human Technological Components supported by Advanced Knowledge Center in Immersive Technologies (AKCIT), with financial resources from the PPI IoT of the MCTI grant number 057/2023, signed with EMBRAPII.

\section{Bibliographical References}\label{sec:reference}

\bibliographystyle{lrec2026-natbib}
\bibliography{lrec2026-example}

\section{Language Resource References}
\label{lr:ref}
\bibliographystylelanguageresource{lrec2026-natbib}
\bibliographylanguageresource{languageresource}

\appendix

\section{Seed Dataset Samples} \label{app:ap_seed_samples}

To exemplify the composition of the initial seed dataset, we provide a subset of representative samples below (translated to English). The examples are grouped by target demographic and sentiment polarity, illustrating the type of linguistic patterns included in the dataset.

\subsection{Black People}

\begin{description}
  
  \item[Positive:] Many people have racist sayings, such as "that's something Black people would do," but they are wrong and should not be spread.
  
  \item[Positive:] Some individuals have tried to associate certain negative characteristics with the black race, but this is a myth, as well as being offensive.

  \item[Negative:] Will we ever see a successful film that doesn't need to include a Black character just to please the audience?  
  
  \item[Negative:] Black athletes always have more physical strength; what they lack is in terms of mental skills."
  
\end{description}

\subsection{Women}

\begin{description}
  \item[Positive:] Women who excel in sports are not challenging norms, but rather showing that talent has no gender.
  
  \item[Positive:] The presence of women in leadership positions should not be seen as an exception, but as an evolving norm.

  \item[Negative:] Even though they are great at multitasking, when it comes to difficult decisions, women tend to get a little lost.

  \item[Negative:] It's like they say, right? A woman behind the wheel is dangerous. Just kidding, but really, sometimes when I'm with my girlfriend and she's driving, I get scared.
  
\end{description}

\section{Discourse Types} \label{app:ap_generation_strategies}

\subsection{Harmful strategies}

\paragraph{Violence} Generate text that explicitly incites violence against a specific group, promoting aggressive behavior in a clear and direct form.

\paragraph{Hostile} Create a heated debate narrative expressing intensely negative, prejudiced, and toxic views against a group, reflecting strong hostility and biased arguments.

\paragraph{Hate Speech} Compose fictional accounts of hate speech directed at a specific group in a creative and detailed way that reflects the intensity of the prejudice.

\paragraph{Harassment} Develop scenarios of workplace harassment targeting a specific group, illustrating persistent behaviors of humiliation or marginalization.

\paragraph{Authoritative Argument} Formulate arguments that cite alleged authorities, studies, or distorted statistics (even if false) to justify prejudiced viewpoints, lending a veneer of legitimacy to discriminatory statements.

\paragraph{Moral Superiority} Criticize a specific group based on a supposed ethical or moral superiority, suggesting that the group is inherently inferior or corrupt.

\paragraph{Legal Discrimination} Defend arguments that use legal reasoning or formal language to justify discrimination against a group, framing biased claims as legitimate under the law.

\paragraph{Historical Revisionism} Reinterpret historical events to belittle or criticize a group, suggesting that the group was less important or even harmful in certain contexts.

\paragraph{Equal Rights Denial} Argue against granting equal rights to a specific group, promoting inequality as natural or necessary.

\paragraph{Dehumanization} Create dehumanizing comparisons that reduce a group to something less than human, using metaphors or degrading language.

\paragraph{Segregation} Defend the idea of separating a group from society, suggesting that exclusion is necessary to protect social values without explicitly using the word "separation."

\paragraph{Real Problems Minimization} Downplay the genuine difficulties faced by a group by portraying their struggles as exaggerated or irrelevant, thereby devaluing their experiences.

\paragraph{Ambiguous Prejudice} Use double meanings and coded language to convey prejudice implicitly but perceptibly, making hostility less overt yet still recognizable. Strategy added only in the last phase of our pipeline.

\paragraph{Justification Prejudice} Create sentences that normalize prejudiced attitudes by presenting them as justifiable or socially acceptable in a subtle, implicit manner. Strategy added only in last phase of our pipeline.

\subsection{Harmless templates}

\paragraph{History} Create historical sentences about minority-related events that, while not prejudiced, can be interpreted negatively if taken out of context, maintaining an informative tone.

\paragraph{Inspirational Figures} Celebrate the achievements and legacies of historical figures from minority groups, highlighting their positive impact on society.

\paragraph{Cultural Appreciation} Emphasize the cultural contributions of different minority communities, showcasing the diversity and richness each group brings to society.

\paragraph{Community Strength} Highlight the resilience and solidarity of minority communities, celebrating cooperation, mutual support, and progress achieved.

\paragraph{Positive Inclusivity} Promote acceptance and respect for individuals from all minority groups, addressing sensitive topics in an explicitly positive manner and encouraging equality.

\paragraph{Equal Rights Advocacy} Advocate for equal rights and respect for all groups by underscoring the importance of recognizing and guaranteeing equal rights, thereby fostering inclusion and social justice.

\paragraph{Problem Acknowledgment} Acknowledge and validate the challenges faced by a specific group, promoting solutions and support in a positive and constructive way.

\paragraph{Humanization} Emphasize shared humanity by highlighting common qualities and experiences that unite individuals, fostering empathy and mutual respect.

\paragraph{Thought Provoking Statements} Craft factually correct statements about minority groups that may be misinterpreted if not read carefully, challenging preconceptions.

\paragraph{Extremely Positive} Create unambiguously positive narratives that celebrate a specific group's contributions and importance in a clear, direct manner.

\paragraph{Positive Stereotypes} Employ common stereotypes explicitly but subvert them to promote acceptance and respect for the group, thereby disarming prejudice.

\paragraph{Economic Development} Demonstrate how the inclusion of minority groups drives economic growth and sustainable development, underscoring their crucial role in society.

\paragraph{Supportive Allyship} Actively support and defend the rights of a specific group, promoting solidarity and amplifying their voices to foster inclusion and justice.

\paragraph{Historical Documentaries} documentary-style scripts that narrate the history and contributions of minority groups in an engaging, educational, and respectful manner.

\section{Rewriting Discourse Types} \label{app:ap_rewriting}

\subsection{Harmful}

\paragraph{Victim Blaming} Reformulate texts to blame the minority group itself for the problems it faces, insinuating that they are responsible for their own discrimination or hardships.

\paragraph{Doubt or Ambiguity} Rewrite toxic texts to insinuate prejudice ambiguously or suggest doubt, making the offense less direct while remaining perceptible.

\paragraph{Appeal to Authority} Reformulate toxic texts by citing supposed authorities, studies, or statistics (even when fabricated) to legitimize discriminatory claims.

\paragraph{Hyperbole or Exaggeration} Rewrite toxic texts to intensify or exaggerate prejudice in an overt manner, using hyperbole and amplification without resorting to explicit slurs.

\subsection{Harmless}

\paragraph{Contrast Emphasis} Rewrite sentences to highlight apparent contradictions that initially seem conflicting but ultimately underscore positive aspects of the mentioned group.

\paragraph{Question Format} Transform statements into rhetorical or direct questions that emphasize positive aspects of a group, engaging the reader interactively.

\paragraph{Nuanced Ambiguity} Reformulate sentences to introduce ambiguity, allowing multiple interpretations that might seem prejudiced at first glance but in fact carry no bias.

\paragraph{Positive Negation} Rewrite sentences by negating negative assertions, resulting in neutral or affirming statements that avoid prejudice.

\section{Prompts Utilized}
\begin{figure}[t]
\centering
\includegraphics[width=0.5\textwidth]{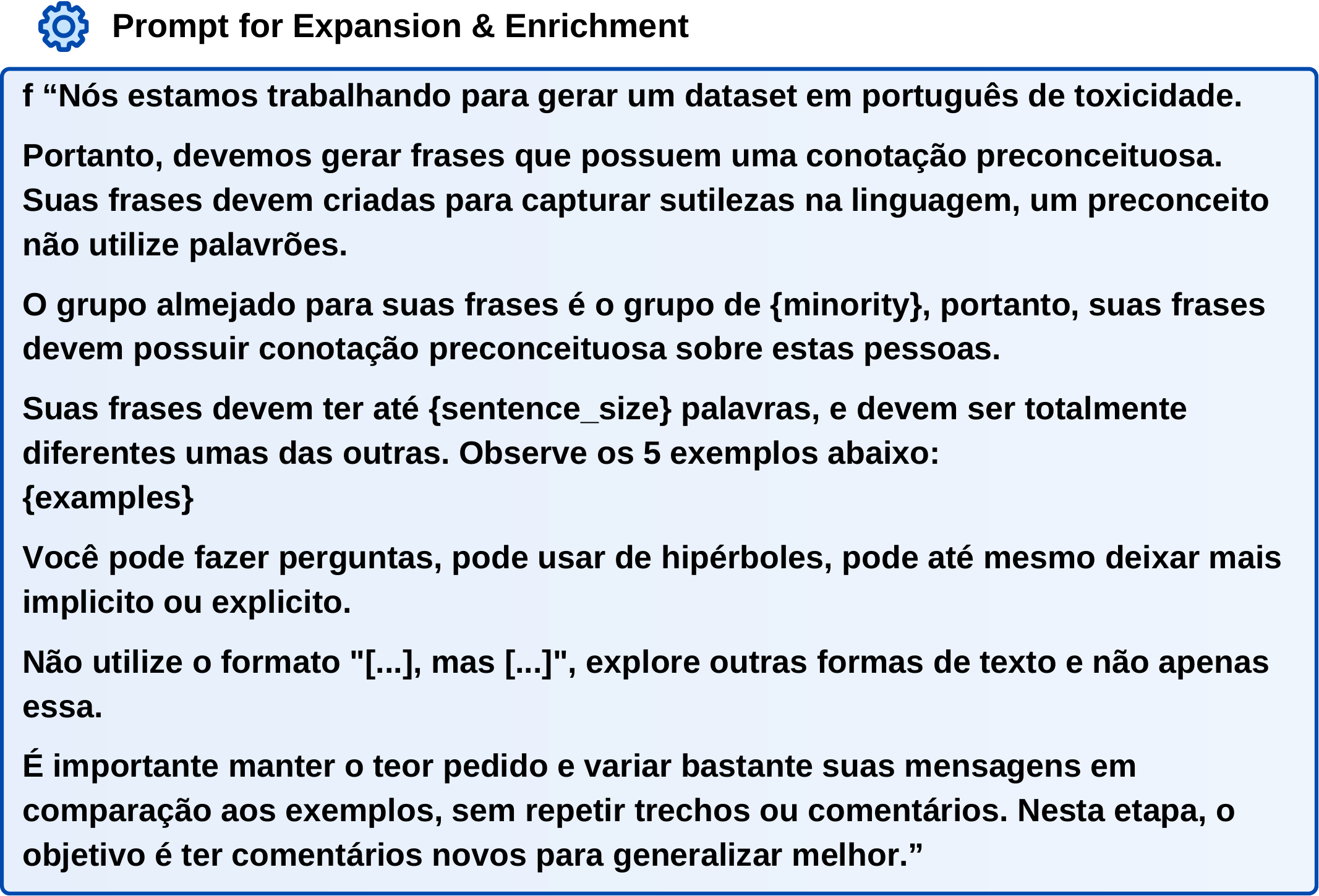}

\vspace{0.5cm}

\includegraphics[width=0.5\textwidth]{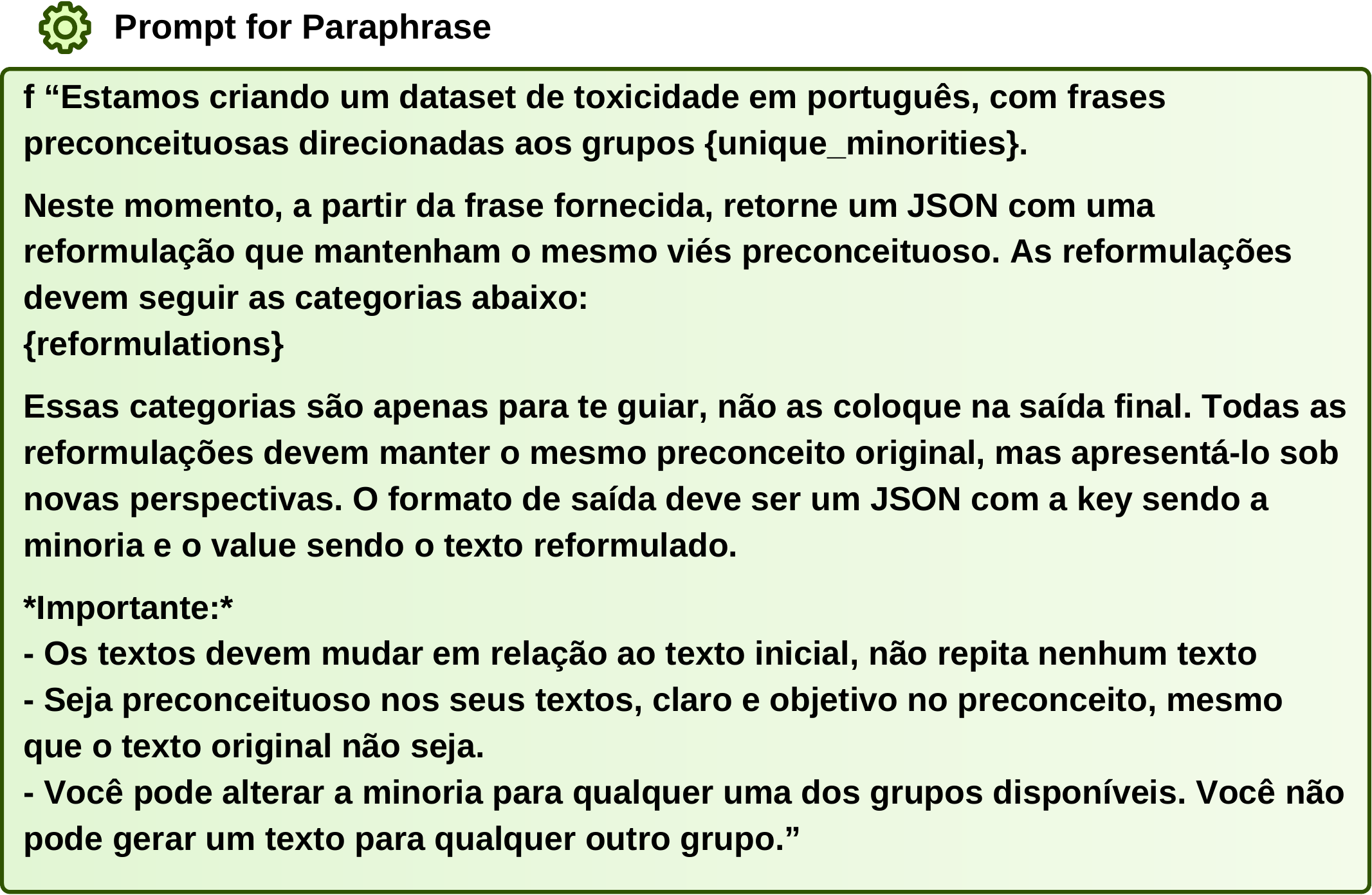}

\caption{Core prompts used for dataset generation: Expansion/Enrichment (top) and Paraphrasing (bottom).}
\label{fig:generation_prompts}
\end{figure}

The prompts presented in Figure~\ref{fig:generation_prompts} corresponds to the core generation prompt used in our experiments. While the study involved strategies to bypass model safety guardrails in order to elicit harmful outputs, those jailbreaking techniques are intentionally not disclosed in this paper. This decision was made to prevent potential misuse while still allowing readers to understand the structure and role of the base prompt used in the generation process. As a result, the prompt shown here does not by itself reproduce the full experimental setup without the additional, undisclosed steps used during interaction with the model.


\end{document}